\documentclass{article}

\PassOptionsToPackage{numbers}{natbib}
\usepackage[nonatbib, final]{nips_2017}

\usepackage[utf8]{inputenc} 
\usepackage[T1]{fontenc}    
\usepackage{hyperref}       
\usepackage{url}            
\usepackage{booktabs}       
\usepackage{amsfonts}       
\usepackage{nicefrac}       
\usepackage{microtype}      
\usepackage{amsmath}
\usepackage{amssymb}

\usepackage{graphicx}
\usepackage{subfig}
\newcommand{\etal}{\textit{et al.} }
\usepackage{color}

\usepackage{pgf}

\definecolor{framecolor}{HTML}{0051A8}
\title{Pedestrian Prediction by Planning \\ using Deep Neural Networks}

\author{Eike Rehder, Florian Wirth, Martin Lauer, and Christoph Stiller \\
		  Institue of Measurement and Control Systems\\
		  Karlsruhe Institute of Technology\\
		  Karlsruhe, Germany\\
		  \texttt{\{eike.rehder, lauer, stiller\}@kit.edu}\\
}

\begin{document}

\maketitle
\begin{abstract}
Accurate traffic participant prediction is the prerequisite for collision avoidance of autonomous vehicles. In this work, we predict pedestrians by emulating their own motion planning. From online observations, we infer a mixture density function for possible destinations. We use this result as the goal states of a planning stage that performs motion prediction based on common behavior patterns. The entire system is modeled as one monolithic neural network and trained via inverse reinforcement learning. Experimental validation on real world data shows the system's ability to predict both, destinations and trajectories accurately.
\end{abstract}
\section{Introduction}

With recent advances in artificial intelligence, the goal of fully automated driving seems to be in reach. However, for safe maneuvering of such vehicles, collisions must be prevented. For its own motion planning, an autonomous vehicle has to predict behavior of other traffic participants. This is especially relevant for vulnerable road users (VRUs), since collisions are likely to be fatal. Thus, accurate prediction of human motion is necessary. 

Prediction of VRUs is often implemented as an extrapolation of observed dynamics, e.g. by recursive Bayesian filtering. See \cite{schneider_pedestrian_2013, keller_will_2014} for detailed studies. For larger lookahead times however, meaningful prediction becomes infeasible due to the highly dynamic nature of human motion. 

To cope with this, classification based prediction has been proposed. These approaches deal with discrete questions, e.g. "Will the pedestrian cross the road?" \cite{bonnin_pedestrian_2014}. They are based on two crucial assumptions for prediction: firstly, pedestrian behavior is intention-driven, and, secondly, this intention can be inferred from more cues than just pure dynamics.

If we specifically address a human's intention, prediction can be modeled as a goal-directed planning task. Here, the intention corresponds to a spacial destination. The works of Ziebart, Kitani \etal proposed Markov Decision Processes (MDPs) together with Inverse Reinforcement Learning (IRL) to solve prediction \cite{ziebart_planning_2009, kitani_activity_2012}. While these works can predict pedestrians over long periods of time, they require static observation and prelocation of possible destinations. Adaptions to dynamic road scenes exist, yet they still employ relatively naive scene models and destination models \cite{rehder_goal_2015, karasev_intent_2016}.

The intention, however, could accurately be inferred from observations. Humans, for example, rely on few visual cues for activity forecasting  \cite{schmidt_pedestrians_2009}. Features such as distance to curb or head orientation have successfully been used for intention recognition in traffic \cite{bonnin_pedestrian_2014, kooij_context_2014}. However, these works neglected the implications of intention inference for planning-based prediction. 

In this work, we propose to solve the problem of intention recognition and planning-based prediction in one single Artificial Neural Network (ANN). The intention in form of destinations is predicted using a recurrent Mixture Density Network (RMDN). We then formulate the planning problem towards these destinations as a Convolutional Neural Network (CNN). We propose two different architectures for planing, namely an MDP network and a forward-backward network. The topology which serves as the basis for these planning algorithms is generated by a fully convolutional network (FCN), operating on maps of the environment. By interconnecting all individual networks, it is possible to train the entire setup jointly in a monolithic inverse reinforcement learning setup.

\section{Prediction Network Architecture}
\begin{figure}
  \begin{center}
    \resizebox{0.92\linewidth}{!}{
     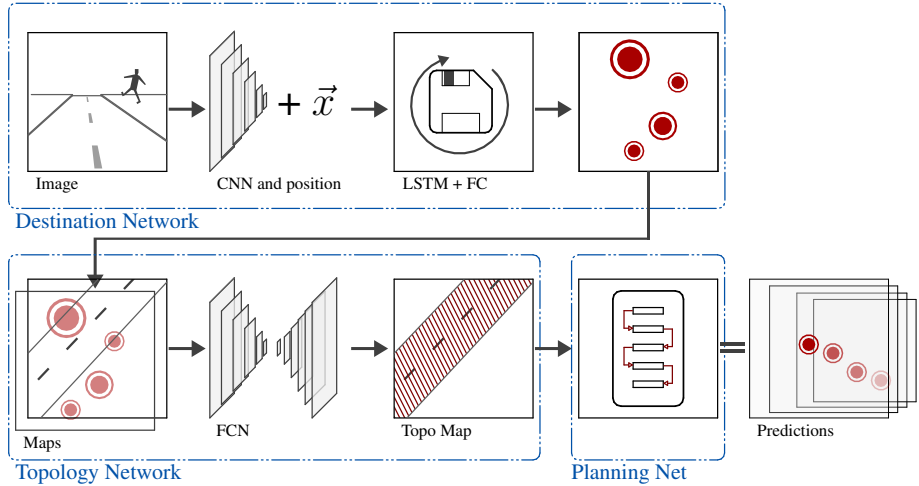
    }
  \end{center}
\caption{Complete prediction network. The Destination Network predicts a mixture of possible destinations from pedestrian images and positions in form of a grid map. The Topology Network generates a planning topology from these destination grid maps as well as other environment features. Finally, the Planning Network runs prediction as planning on these topology maps. The final output is a position probability map per predicted time step.}\label{fig:completenet}
\end{figure}

The proposed network combines two major ideas: firstly, we infer a pedestrian's destination from images and positions. Secondly, we apply trajectory planning towards these destinations for prediction. For this, the network is augmented from three individual parts as depicted in Figure \ref{fig:completenet}. The first part consists of a recurrent mixture density network (RMDN).  Images and position of the pedestrian serve as an input. The image is processed with a standard CNN the input to the LSTM is then a concatenation of CNN output and position vector. As output, the network predicts possible destinations in form of a probability distribution map.

This map is fed to the second part of the network, the Topology Network. Here an FCN predicts a map on which planning will be executed. In the context of graph solvers, this may be understood as a cost map. However, depending on the planning algorithm that is applied, it may take the form of a reward or a state probability. The planning is executed on the topological map to yield the actual prediction. In this work, we show how to incorporate two different planning techniques into our model, namely Markov Decision Processes and the Forward-Backward Algorithm. The final prediction output is a set of one position probability grid per predicted time step.

\section{Destination Prediction}\label{sec:destination_prediction}

Destinations are modeled as a mixture density function. The intuition behind this is that every component represents one destination alternative. The underlying density function then models the uncertainty within one specific alternative. Our destination model consists of a mixture of Gaussian-von-Mises distributions with mean position $\vec{\mu}_i {=} (\mu_{x,i}, \mu_{y,i})^{\top}$ and covariance matrix $\Sigma_i$ as well as mean heading angle $\gamma_i$ and concentration $\kappa_i$. Mixing coefficients $\pi_i$ weight the individual components. 

The final probability density function is given as 
\begin{eqnarray}
p(\vec{x}, \psi) {=} \sum_{i=1}^{N} \pi_i 
  \underbrace{
    \frac{1}{\sqrt{|2\pi\Sigma_i|}} 
    \exp\left({-}\frac{1}{2}(\vec{x} {-} \vec{\mu}_i)^{\top} \Sigma_i^{-1} (\vec{x} {-} \vec{\mu}_i)\right)
  }_{\text{general bivariate Normal Distribution for position}}
  \underbrace{
    \frac{1}{2\pi I_0(\kappa_i)}\exp(\kappa_i\cos(\psi {-} \gamma_i)) 
  }_{\text{von-Mises-Distribution for heading}}. \label{eq:mixturecomp}
\end{eqnarray}

To estimate the mixture from observations, we infer $\vec{\mu}_i$, $\Sigma_i$, $\gamma_i$, $\kappa_i$, and $\pi_i$. Mixture Density Networks (MDNs) have been proposed to solve this task \cite{bishop_mixture_1994}. Outputs of a neural network are adjusted with different activation functions to satisfy the constraints of probability distributions. 

One mixture component is made up from a set of neuron outputs $\{m_x, m_y, s_x, s_y, r, p, k, g\}_i$, where $i {\in} 1,\ldots,N$ for $N$ mixture components. The Gaussian part is then computed with the neuron activations $\sigma_{x,i} {=} \exp(s_{x,i})$, $\sigma_{y,i} {=} \exp(s_{y,i})$, and $\rho_i {=} \mathrm{tanh}(r_i)$, so that
\begin{eqnarray}
\vec{\mu}_i {=} (m_{x,i}, m_{y,i})^{\top}\text{\hspace{.5cm} and \hspace{.5cm}}
\Sigma_i {=} \left[\begin{matrix}
\sigma_{x,i}^2 & \rho_i\sigma_{x,i}\sigma_{y,i}\\
\rho_i\sigma_{x,i}\sigma_{y,i} & \sigma_{y,i}^2\\
\end{matrix}\right]
\end{eqnarray}
to satisfy the constraints of the covariance matrix. Equivalently, the von-Mises part in \eqref{eq:mixturecomp} can be generated from the input as $\gamma_i {=} g_i$ and  $\kappa_i {=} \exp({k_i})$, so that $\kappa_i {>} 0$ at all times. Finally, the mixing coefficients are  the softmax over all corresponding network outputs of all mixture components $\pi_i {=} \exp(p_i)/\sum_{j=1}^{N}\exp(p_j)$.

The MDN requires an output of the size $8 \times N$ for eight parameters of $N$ mixture components. Prediction is run on time series of pedestrian tracks. We use a recurrent MDN (RMDN) with a long short-term memory (LSTM) cell for time series processing \cite{sak_long_2014}. As an input we use both, pedestrian images and positions.



\section{Planning Network}\label{sec:planning}
Prediction is modeled as reenactment of goal-directed motion planning. Thus, a suitable model for rational motion planning has to be found. In robotics, it is common to execute planning algorithms in a discretized state space, i.e. grid maps or state lattices. With such a structure, a motion sequence can be interpreted as a graph and be solved efficiently. 

When the state space is a grid map and the action radius is limited, the transition function can be modeled as iterative convolution applied to the state grid \cite{rehder_goal_2015, tamar_value_2016, shankar_reinforcement_2017}. 

We assume a state grid of discrete states at time $t$, $S_t {=} \{s_1, \ldots, s_N\}_t$, and a set of $M$ actions $A {=} \{a_1, \ldots, a_M\}$. With this, the probability of ending up in a specific state of $S_{t+1}$ can be expressed as discrete convolution of the state grid with the action-dependent transition filter mask $P_{a_i}(S_{t+1}| S_{t})$  
\begin{eqnarray}
P(S_{t+1} | a_i) {=} P_{a_i}(S_{t+1}| S_{t})\otimes P(S_{t}).\label{eq:conv_mdp}
\end{eqnarray}

The problem of motion planning then is the selection of appropriate actions $a_i$  for every state in $S_t$ in order to reach ones goal. 

The full prediction problem thus consists of three individual parts: firstly, we incorporate the prediction of destination as presented in Section \ref{sec:destination_prediction} into a grid structure. Next, we find a policy function that selects appropriate actions per state and time. In the third part, we simulate a person's motion in the world according to the derived policy to obtain the actual prediction result.

\subsection{Markov Decision Process Network}
The selection of optimal actions in the presence of uncertain transition functions is known as a Markov Decision Process (MDP). In this context, one tries to find a policy to select optimal actions per state in order to maximize the expected reward of successive actions. 

In our context, this means that the reward is linked to the arrival at the destination and the policy represents the steps to take to reach those destinations. The training of policy,  transition models and reward function from observed trajectories is referred to as Inverse Reinforcement Learning (IRL) or Imitation Learning (IL) since the algorithm should learn to imitate human behavior \cite{ziebart_planning_2009, kitani_activity_2012, karasev_intent_2016}.

One solution to this problem is Value Iteration (VI) 
\begin{alignat}{3}
V_{k+1}(s) &{=} \max_{a_i\in A}\sum_{s'} P_{a_i}(s, s')(R_{a_i}(s,s') {+} \gamma V_k(s')), \label{eq:value_funcion}\\
\pi_{k+1}(s) &{=}\arg \max_{a_i \in A} \sum_{s'} P_{a_i}(s, s')(R_{a_i}(s,s') {+} \gamma V_k(s')), \label{eq:policy_function} 
\end{alignat}
where the value $V_k(s)$ represents the currently expected total reward and $R_{a_i}(s,s')$ is the reward of a state transition from $s'$ to $s$ by choosing action $a_i$. The constant $\gamma {\in} [0,1]$ is a discount factor that leads to the preference of early rewards. With VI, a value function is found by iteratively applying \eqref{eq:value_funcion} until convergence. The policy function  $\pi(s)$ then is the argmax over all possible actions of the expected reward.
Since only one action is selected in every state, Eq. \eqref{eq:conv_mdp} for prediction becomes
\begin{eqnarray}
P(S_{t+1}) {=} \sum_{s_i \in S_t}P(S_{t+1}| s_{i}, a_i)P(s_{i})\pi(s_i). \label{eq:mdp}
\end{eqnarray}
Recently, Tamar \etal as well as Shankar \etal have proposed to solve this algorithm with simple CNN techniques \cite{tamar_value_2016, shankar_reinforcement_2017}.

\begin{figure}
  \begin{center}
    \resizebox{0.95\linewidth}{!}{
      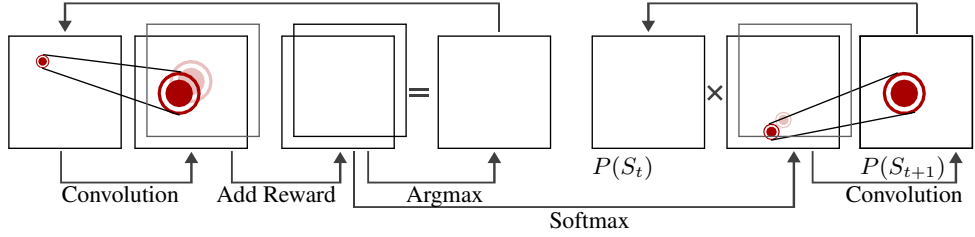
    }
  \end{center}
\caption{MDP-Network. The left part performs Value Iteration to find the policy, the right part simulates motion according to that policy.}\label{fig:mdp}
\end{figure}

Figure \ref{fig:mdp} shows the CNN realization of the MDP. The left part recursively computes \eqref{eq:value_funcion} with a convolution of the current value map with all transition filters. The result is one new value map per action. To this the reward is added and max pooling is performed over the action channel. Upon convergence, the per-action value map is softmaxed to derive a policy map. This policy map is multiplied with a current state grid to select transitions per state. By convolution with the transition filters, probability propagation can be computed recursively for prediction.

\subsection{Forward-Backward Network}

The MDP takes the assumption that an agent always selects the optimal policy to reach the goal and that the only uncertainty is introduced by the state transitions. While this may be true for automated control, it may not be true for human motion. Specifically, a human may choose suboptimal actions at random since human motion planning is no real optimization system. To model even sub-optimal paths towards a known destination, the Forward-Backward Algorithm can be applied. It computes the probability of all intermediate states,
\begin{alignat}{3}
P( & S_{t}) {\propto} \nonumber \\& \underbrace{\sum_{a_i \in A} \left(P_{a_i}(S_{t} | S_{t-1}){\otimes} P(S_{t-1}) {\cdot} P(a_i | S_{t-1} )\right)}_{\text{forward propagation}} {\cdot} \underbrace{\sum_{a_i \in A}\left(P_{a_i}(S_{t}| S_{t+1}){\otimes} P(S_{t+1}) {\cdot} P(a_i | S_{t+1})\right)}_{\text{backward propagation}}.\label{eq:fwdbwd}
\end{alignat}
For this, it propagates the probabilities from the starting state to the future and from the goal state to the past. The product of the two gives the joint probability \cite{rehder_goal_2015}. The probabilities $P(a_i | S_{t-1})$ and $P(a_i | S_{t+1})$ represent the distribution over actions to be taken in specific time steps.

\begin{figure}
\begin{center}
  \resizebox{0.95\linewidth}{!}{
    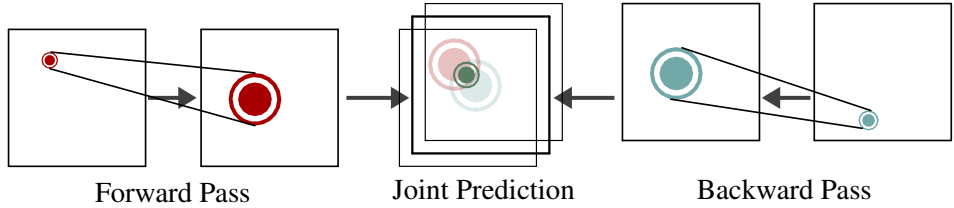
  }
\end{center}
\caption{Forward-Backward-Architecture. The fwd pass models motion from start to every state in the grid, the bwd pass models motion from every state towards the destination. The product of the two represents the transition step between start and destination.}\label{fig:fwdbwd}
\end{figure}

The network architecture of \eqref{eq:fwdbwd} is depicted in Fig. \ref{fig:fwdbwd}. The forward pass is computed as recursive convolution of the starting state grid with the transition filters. The backward pass is equivalently computed from the destination state grid with flipped transition masks. Multiplication of the time-step-wise results yields the predicted trajectory.

\subsection{Destination Mapping}
In Section \ref{sec:destination_prediction}, we presented an RMDN to predict distributions for pedestrian destinations in continuous domain. However, since we employ the planning methods in a discrete state space, we need to discretize the output of our RMDN. For this, we apply mesh grids for all possible values of $\vec{x}$ and $\psi$ in Eq.  \eqref{eq:mixturecomp} as a constant additive layer in the arguments of the RMDN. The result is a three dimensional grid with discretization of positions in $x$- and $y$-direction and discrete orientation angles in depth. In the output of the RMDN, each cell represents how likely it is that a pedestrian is in the corresponding position and orientation state at some prediction time.

\subsection{Topology Network}
The motion planning requires knowledge about the environment to derive the action selection policy. In the MDP setting, it is represented as the reward of states. In the Fwd-Bwd setting, it governs the probability of actions in specific states. In any case however, it represents a topological map that connects behavior to locations.

The TopologyNet is dependent on two different inputs, namely the destination to which a person is headed and the features of the surrounding, e.g. obstacles. In order to be adoptable to a great variety of situations, we employ Inverse Reinforcement Learning (IRL) to train an FCN that predicts the policy from the RMDN destination as well as environmental features. 

We train a reward function as an FCN with the predicted destination as well as on features of the environment as input. When the output is the reward for the MDP, we do not apply any non-linearity to the output of the network. For the Fwd-Bwd-Net, since the outputs should represent the transition probabilities, we apply sigmoidal activation and normalize later.

\section{Network Training Details}\label{sec:training}

The full prediction network is augmented from the three individual networks. To train this joint setup, some considerations on training setups have to be taken into account.

\subsection{Destination Mixture Density Network}

The RMDN consists of the image processing part and the recurrent LSTM part. It is trained from manually annotated time series of pedestrian tracks. In order to achieve stable results, we introduced the following tweaks to the training process.

\paragraph{Function Modeling}
The von-Mises part of Eq. \eqref{eq:mixturecomp} requires a $\kappa_i$-dependent normalization factor in form of the modified Bessel function $I_0(\kappa_i)$. We use an accurate Taylor approximation for normalization estimation.

\paragraph{Training Loss}
The network can readily be trained with the negative log likelihood function of the output evaluated in ground truth destinations as a loss function. 

For the $k$th training example, the loss is given from \eqref{eq:mixturecomp} evaluated in the ground truth (GT) as $l_k {=} {-}\log(p(\vec{x}_{GT,k}, \psi_{GT,k})).$ One could train the network with the average of per-example loss per training batch of size $M$. However, we found it beneficial to use the minimum over the batch as $l {=} \min(l_1, \ldots, l_{M})$. 

Furthermore, we apply dropout on the mixing coefficients $\pi$ to randomly disable some of the mixture coefficients. These two measures are necessary to ensure meaningful training of all mixture components.



%

\subsection{Planning-Networks}

We train all parameters of the planning jointly. That also includes transition filters per action with no special treatment as in \cite{shankar_reinforcement_2017} where knowledge over the selected action was required.

\paragraph{Filter Masks} 
The transition filter masks have to satisfy the constraints of a probability distribution. For this, each filter element $f_{a,i}$ for action $a$ is made up from a set of weights $w_{a,i} \in W_a$ as $f_{a,i} {=} \exp(w_{a,i})/\sum_{w_{a,j} \in W_a}\exp(w_{a,j})$. As initialization, we sample the weights from a standard normal distribution for greater variety and smooth the result using a box filter. Also, we regularize the transition filters pairwise to be complementary. This is done by regularization of Frobenius inner product.
%

\paragraph{Training Loss}
We minimize the negative log likelihood of predictions evaluated in ground truth pedestrian positions.

 
%
%
%
%

\subsection{Joint Network}

For small time horizons, we found that just the prediction loss is sufficient to train the joint net. However, for larger horizons, the destination network does not converge, rendering the entire predictor futile. Thus, the joint network is trained with a compound loss from all individual loss functions of all network parts. Specifically, we add up the log likelihood loss from the RMDN and the planning network as well as the weighted regularization loss.

\section{Experiments}\label{sec:experiments}

The proposed network is trained and evaluated on real world data. For this, we collected stereo videos from multiple drives through urban and residential areas. In the videos, we manually annotated all pedestrians. From stereo imaging, we optimized pedestrian trajectories with a constant velocity motion model in offline processing. In total, we ended up with roughly 400 individual pedestrian tracks that were split into training and test data. The data will be made publicly available as subject of another publication. All results shown here were evaluated on the independent test set.

\begin{figure}
\subfloat[][Image from vehicle-borne camera taken in an urban area. The paths of the two framed example pedestrians are shown in (b)]{
  \includegraphics[width=.48\linewidth]{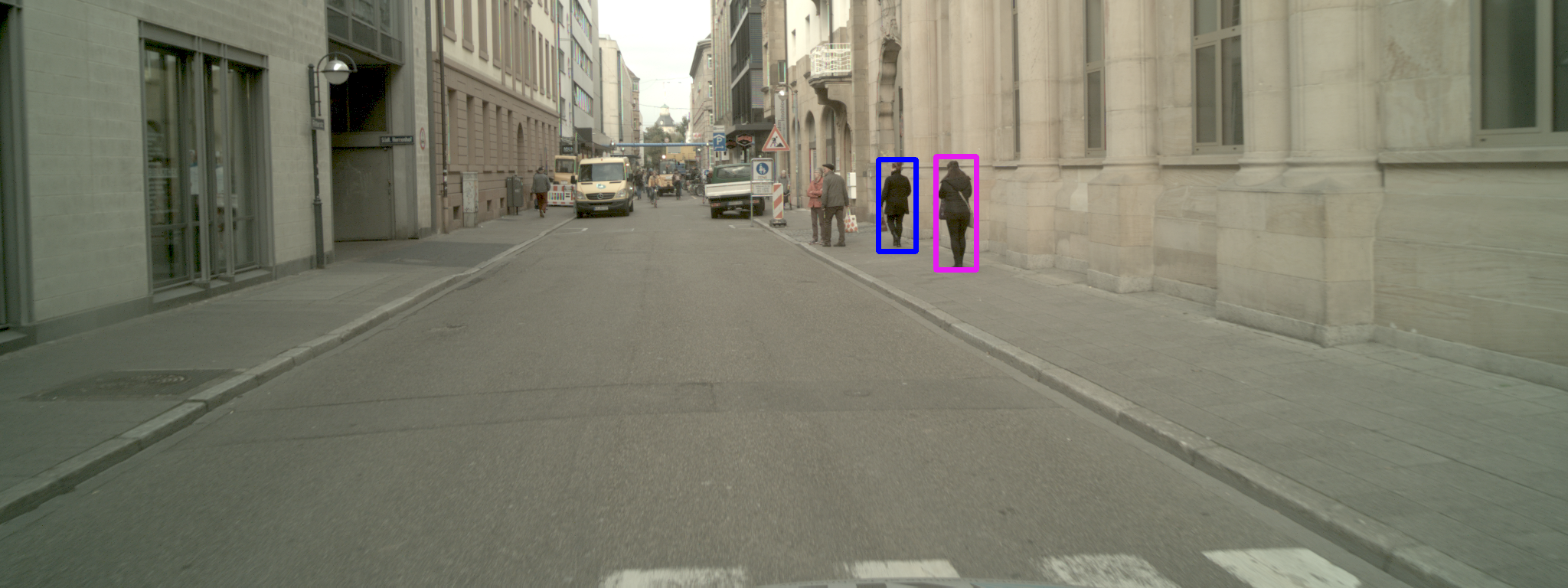}
}
\hfill
\subfloat[][Corresponding topview with feature grids around the vehicle (black box). Mapped features are road (cyan), sidewalk (yellow) and obstacles (red)]{
  \includegraphics[width=.48\linewidth]{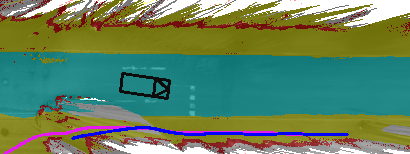}
}
\caption{Input data for prediction: (a) camera image as input for destination inference and (b) feature map for planning. Two pedestrians with their paths are shown for reference.}\label{fig:data}
\end{figure}

For features of the environment, we construct a set of occupancy grid maps \cite{elfes_using_1989}. We segment the camera image into semantic features with a fully-convolutional GoogleNet trained on the Cityscapes dataset \cite{szegedy_going_2015, long_fully_2015, cordts_cityscapes_2016}. By the means of stereo vision \cite{ranft_modeling_2014}, we map the features \emph{obstacles}, \emph{road}, \emph{sidewalk} and \emph{topview}. Fig. \ref{fig:data} shows the camera image of a road scene together wit the corresponding maps.

As performance measure for our prediction, we evaluate the predicted probability distribution in small area around the ground truth (GT) observation. The area is a circle of roughly $0.1m^2$ centered around GT positions. For reference, we also evaluate prediction performance of a constant position-constant velocity Interacting Multiple Model Filter (IMM) \cite{schneider_pedestrian_2013}. Covariance matrices were determined from GT process variation and observation noise was set to $10^{-8}$. 




\subsection{Destination Prediction}

In a first experiment, we evaluate pure destination prediction. Additionally to the IMM reference, we also tested a recurrent density network (RDN), effectively predicting \eqref{eq:mixturecomp} with a single mixture component. Our RMDN features eight mixture components.


\begin{figure}
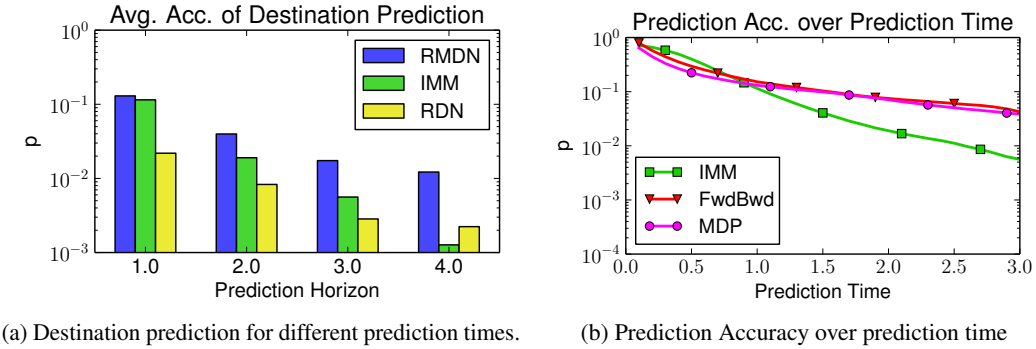

\begin{center}

\subfloat[][Destination prediction for different prediction times. ]{
  \resizebox{0.48\linewidth}{!}{
  \input{dest_pred_time.pgf}
  }\label{fig:destpredtime}
}
\hfill
\subfloat[][Prediction Accuracy over prediction time]{
  \resizebox{0.48\linewidth}{!}{
  \input{pred_over_time.pgf}
  }\label{fig:trajpredtime}
}
\end{center}
\caption{Results of destination prediction}\label{fig:eval}
\end{figure}

Fig. \ref{fig:destpredtime} shows the average predicted probability of GT destinations for different prediction horizons. Since the process noise in the IMM has to represent all uncertainty in dynamics, the covariance spreads greatly. The RDN predicts all possible motion outcomes in a single distribution, thus is even less accurate. Only the RMDN can make full use of all available information and accurately represent different motion patterns. It is noteworthy that while the recurrent filter accuracy falls of linearly over prediction time, the decrease of the RMDN accuracy appears to be sub-linear.


\subsection{Trajectory Prediction}

In a second experiment, we evaluate the entire prediction network on real world data. We run the RMDN on tracks of pedestrians of three seconds length and predict three seconds ahead for each time step. The evaluation method is the same as above. 


Figure \ref{fig:trajpredtime} shows the results of full prediction evaluation. As expected, the prediction certainty decreases over predicted time step. However, the decline is much slower than that of the IMM. This is due to the fact that the IMM has to model all uncertainty about the path in the process noise. Our model on the other hand can make use of much more certain destination predictions as well as the corresponding planning. However, our model uses fixed transition distributions per step. These of course have to model motion uncertainty and are limited to the cell discretization. Thus, for the first second, the spread of the distribution is larger than that of the IMM. 

The MDP network performs slightly better than the Fwd-Bwd network. However, results are very similar. In our experiments, we found that destination forecasting is the crucial influence for our prediction models. 


\subsection{Example Prediction}

\begin{figure}
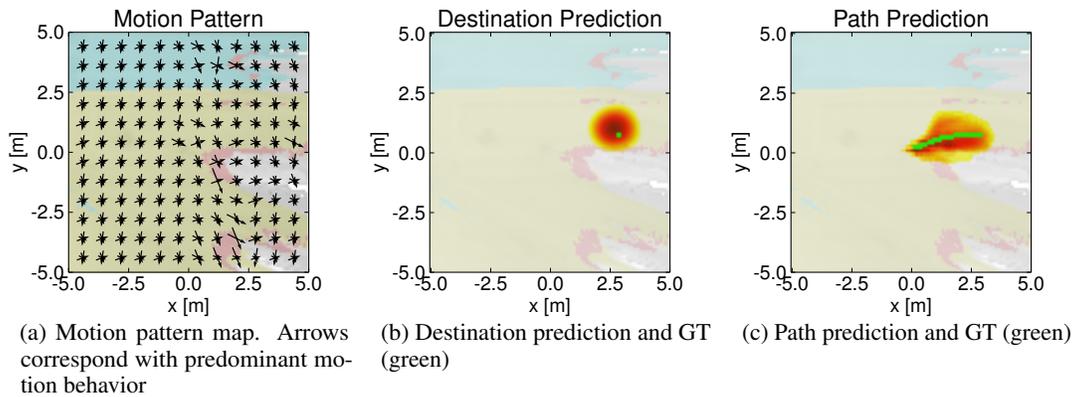

\begin{center}
\subfloat[][Motion pattern map. Arrows correspond with predominant motion behavior]{
  \resizebox{0.3\linewidth}{!}{
  \input{res_motion_behave_2.pgf}
  } \label{fig:motionpattern}
}
\hfill
\subfloat[][Destination prediction and GT (green)]{
  \resizebox{0.3\linewidth}{!}{
  \input{res_dest_pred_2.pgf}
  } \label{fig:destprediction}
}
\hfill
\subfloat[][Path prediction and GT (green)]{
  \resizebox{0.3\linewidth}{!}{
  \input{res_path_pred_2.pgf}
  } \label{fig:pathprediction}
}
\end{center}
\caption{Results of prediction. For reference, a crop of the feature map is also displayed} \label{fig:example}
\end{figure}

Figure \ref{fig:example} shows one example prediction using the Fwd-Bwd setting run on the scene as depicted in Fig. \ref{fig:data}. The pedestrian is located in the center of the map. 

In \ref{fig:motionpattern}, the most likely action directions as predicted by the topology network are displayed as vectors. Note how the vectors display a tendency to avoid obstacles. Figure \ref{fig:destprediction} shows the predicted destination. Here, all mixture components coincide in one location. 

Finally, the predicted path is displayed in Fig. \ref{fig:pathprediction}. Note how the maximum of the predicted probability distribution avoids the obstacle nicely. This kind of behavior could not be modeled with a standard recursive dynamic filter approach.
 
\section{Conclusion}

In this work, we proposed the use of planning for prediction in a monolithic neural network architecture. By prediction of possible destinations, planning towards these can be used as a predictor. 


To the best of our knowledge, this is not only the first work that solves destination forecasting for planning-based prediction. It is also the first work to solve both problems jointly and on the basis of both, image and map information. 

In this work, we relied on precomputed map features. With more training data, however, it may be possible to train the models exclusively with obstacle and top view maps that can be recorded with no manual annotation at all. The implication of this is considerable: an autonomous vehicle that is equipped with means of pedestrian recognition may be able to train the entire pipeline from online observations. This way, even more abstract environment features could be understood by the system, e.g. road markings, cross-walks or traffic refuges. 

\subsubsection*{Acknowledgements}
The research leading to these results has received funding from the German collaborative research center ``SPP 1835 - Cooperative Interacting Automobiles'' (CoInCar) granted by the German Research Foundation (DFG).
\bibliographystyle{splncs}
\bibliography{references}

\end{document}